\documentclass[3p]{elsarticle}

\usepackage{lineno,hyperref}
\usepackage{graphicx}
\usepackage{adjustbox}
\usepackage[table]{xcolor}
\usepackage{multicol}
\usepackage{multirow}
\usepackage{arydshln}
\usepackage{pgfplots}
\usepackage{pgfplotstable}
\usepackage{sfmath}
\usetikzlibrary{patterns}
\usepackage{soul}
\usepackage{float}
\usepackage{subfloat}
\usepackage{subfig}
\usepackage{amsmath}
\usepackage{amssymb}
\usepackage{rotating}
\usepackage{pdflscape}
\usepackage[ruled,vlined]{algorithm2e}
\usepackage{tcolorbox}

\modulolinenumbers[5]

\journal{ArXiv}

\begin{document}

\begin{frontmatter}

\title{
FRAKE: Fusional Real-time Automatic Keyword Extraction\\
}

\author[cominsys]{Aidin Zehtab-Salmasi}\ead{a.zehtab97@ms.tabrizu.ac.ir}
\author[cominsys]{Mohammad-Reza Feizi-Derakhshi\corref{cor}}\ead{mfeizi@tabrizu.ac.ir}

\author[UT]{Mohamad-Ali Balafar}\ead{balafarila@tabrizu.ac.ir}

\cortext[cor]{mfeizi@tabrizu.ac.ir}

\address[cominsys]{Computerized Intelligence Systems Laboratory, Department of Computer Engineering, University of Tabriz, Tabriz, IRAN.}
\address[UT]{Department of Computer Engineering, University of Tabriz, Tabriz, IRAN.}

\begin{abstract}
{Keyword extraction is the process of identifying the words or phrases that express the main concepts of text to the best of one's ability. Electronic infrastructure creates a considerable amount of text every day and at all times. This massive volume of documents makes it practically impossible for human resources to study and manage them. Nevertheless, the need for these documents to be accessed efficiently and effectively is evident in numerous purposes. A blog, news article, or technical note is considered a relatively long text since the reader aims to learn the subject based on keywords or topics. Our approach consists of a combination of two models: graph centrality features and textural features. The proposed method has been used to extract the best keyword among the candidate keywords with an optimal combination of graph centralities, such as degree, betweenness, eigenvector, closeness centrality and etc, and textural, such as Casing, Term position, Term frequency normalization, Term different sentence, Part Of Speech tagging. There have also been attempts to distinguish keywords from candidate phrases and consider them on separate keywords. For evaluating the proposed method, seven datasets were used: Semeval2010, SemEval2017, Inspec, fao30, Thesis100, pak2018, and Wikinews, with results reported as Precision, Recall, and F- measure. Our proposed method performed much better in terms of evaluation metrics in all reviewed datasets compared with available methods in literature. An approximate 16.9\% increase was witnessed in F-score metric and this was much more for the Inspec in English datasets and WikiNews in forgone languages.}

\end{abstract}


\end{frontmatter}


\section{Introduction} \label{sec:Introduction}
These days, people have difficulty finding a particular document among the vast volume of documents they create daily. There are typically millions of text and web pages stored every day, making them impossible to analyze without indexing. Accordingly, if the documents index with expressions, analyzing them becomes highly convenient. Using an expression makes understanding a document much more accessible. An expression can be a single or a series of words or phrases called keywords or key phrases. Keyword or Key-phrase extraction is the process of identifying the primary concept of a document by examining the set of terms composing it\cite{berry_kogan_2010}.

There are three approaches to keyword or phrase extraction in natural language processing: text approach, graph approach, and hybrid. Text-based approaches based on textual features could be used in extracting text from documents, such as Part of Speech (POS), mean TF, etc. The graph modeling approach is to represent words and relations as nodes and edges in a graph by co-occurrences or N-grams. When graphs are created, statistical analysis is applied to scoring nodes to select the top words as keywords. Hybrid models rely on both text and graph methods in order to extract keyword (phrase) combinations.

The rest of this study is organized as follows: the second section contains a literature review of mobile price prediction. Section \ref{sec:proposed_method} presents the proposed methods. The experimental results are given and discussed in section \ref{sec:experiments_results}. Some conclusions are drawn in the final section, and the areas for further researches are identified, as well.
\section{Related Works} \label{sec:related_works}
A two-pronged approach may apply to studies relating to keywords: Extraction \& Generation.This paper employs a keyword extraction approach, meaning that the keywords should appear in the document to be selected as keywords. Based on the second stage of the division of keyword extraction methods, related words can be sorted into four groups: statistical, textual, graph-based, and hybrid. In statistical methods, top-scoring words were chosen as index words of the document, and the score was calculated arithmetically. TF-IDF\cite{Lott2012} and co-occurrence\cite{matsuo2004keyword} can be listed as the most well-known statistical keywords extractor methods.

Textual methods are based on the linguistic features of words in a document, such as lexical, syntactic, and semantic. Part of speech tags of words of a document was one of the text-based keyword extraction methods that ignored stop words\cite{10.3115/1119355.1119383}. Morphology is a linguistic analysis subset that is used in the keyword extraction process. Morphology-based approaches\cite{li2015keyphrase} need a thesaurus, and WordNet\cite{miller1998wordnet} is the most famous one. However, language dependency is the major drawback of this approach. Additionally, some linguistic characteristics in textual-based keyword extraction can be described as "Noun phrase chunking"\cite{hulth2003improved}, Ontology\cite{shamsfard2008} or "Lexical chains"\cite{enss2006investigation}.

The idea behind graph-based methods is to construct a graph from document elements. Word co-occurrence networks are often used to show the relationships between words in a document. There is an edge between two nodes if the relevant words co-occur within a window in a co-occurrence graph with nodes representing words. Additionally, most centrality metrics such as degree, closeness, betweenness, and eigenvector are used to identify the nodes with the highest score, and their keywords are then identified as candidates. Graph-based methods have gained good results in this area, and several methods exist; one of them is Text Rank.\cite{Mihalcea2004}, Single Rank \cite{Wan2008}, Topic page Rank\cite{Sterckx2015}, Position Rank\cite{Florescu2017}, Multipartite Rank\cite{Boudin2018}, Expended Rank\cite{Wan2008}.

These hybrid models combine two previously mentioned categories, textual-based and graph-based. In a hybrid model, the aim is to calculate each text and graph-based feature separately and then combine them in a method. Combining and scoring methods are the main contribution of works. Mike\cite{10.1145/3132847.3132956}, Sgrank\cite{sgrank-2015}, Key2Vec\cite{mahata-key2vec}, RaKUn\cite{rakun2019} are examples of hybrid methods.
\section{FRAKE} \label{sec:proposed_method}
\begin{figure}
    \centering
    \includegraphics[width=\textwidth]{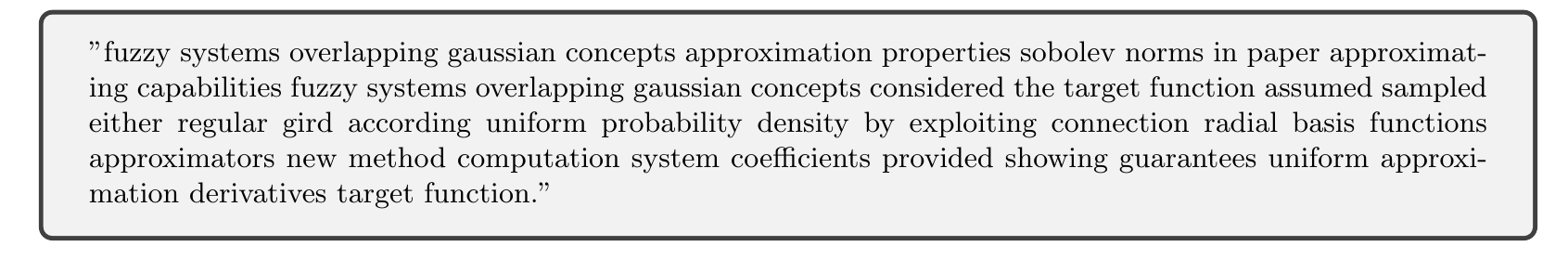}
    \caption{Sample document.}
    \label{fig:sample_1}
\end{figure}

\begin{figure}
    \centering
    \includegraphics[width=\textwidth]{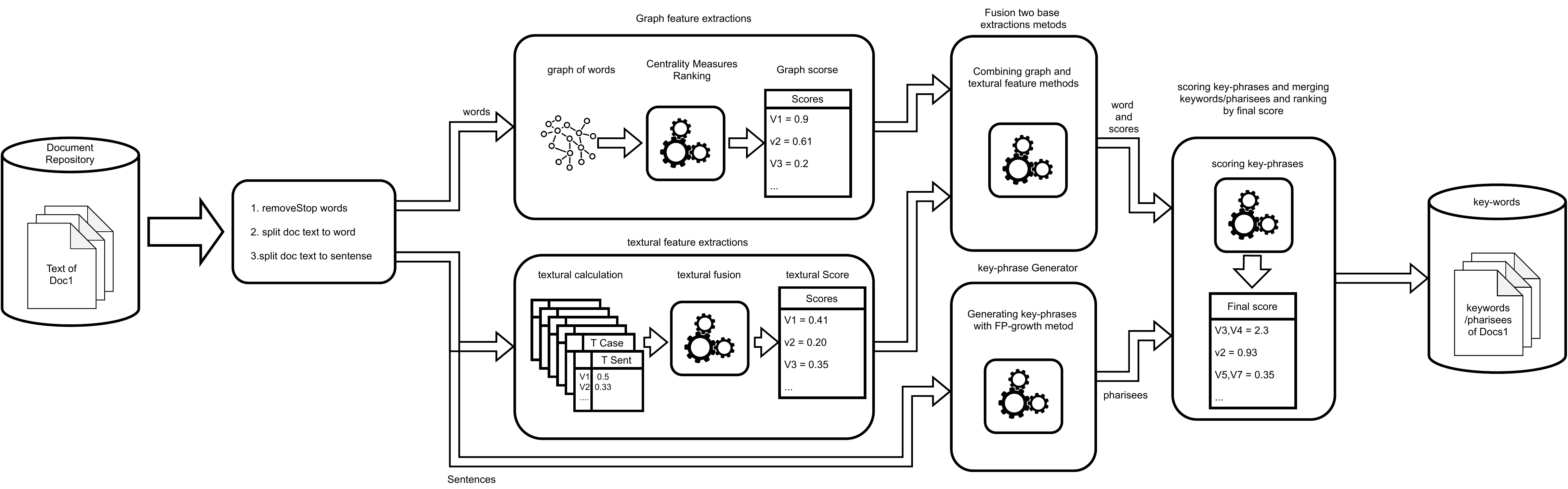}
    \caption{Diagram of the proposed FRAKE model.}
    \label{fig:diagram}
\end{figure}

The proposed method, called FREAK, is a fusion of two parallel keyword extraction approaches, graph features and textural features, and each of the techniques has its advantages. Illustration of the proposed method is shown in fig. \ref{fig:diagram}. The proposed method consists of 5 steps; pre-processing, feature extraction, scores computation, key-phrase generation, and ranking, respectively, with the aim of extracting keywords. 
In the following sub-sections, every step is expanded. To make each step sensible, the output of those steps are shown with an example. The example document is shown in fig \ref{fig:sample_1}.

\subsection{Pre-processing}
\begin{figure}
    \centering
    \includegraphics[width=\textwidth]{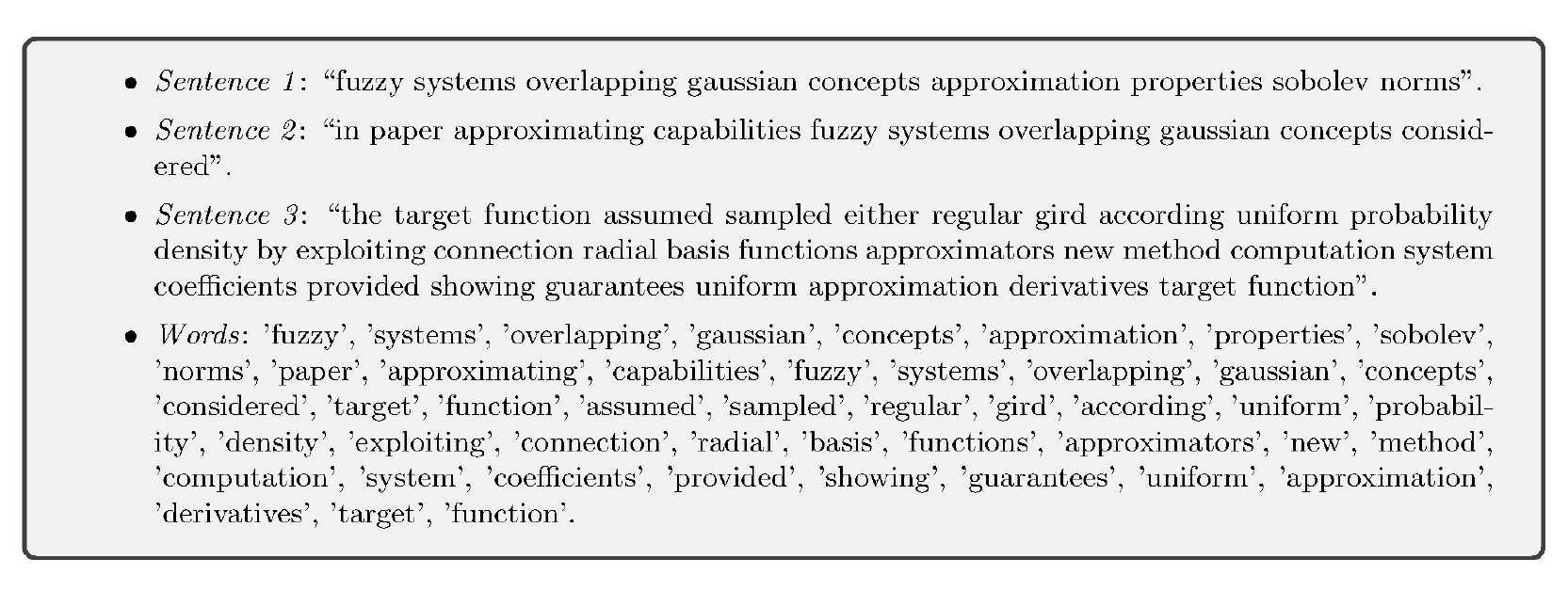}
    \caption{The sentence segmentation and tokenization of example document fig \ref{fig:sample_1}.}
    \label{fig:sample_2}
\end{figure}

As discussed at the beginning of the section, our method consists of two parallel stages in the primary stage, one of which is for the first stage and the second part for the second stage. To proceed with extracting keywords from the graph, we need the word list and the relationship of all the text words with their neighbors. As opposed to the extraction of keywords from the local benchmark, we need to separate the sentences. During this step, the input data are separated into words, sentences, and stop words are dropped from the input data. The example output is shown in fig \ref{fig:sample_2}.

\begin{algorithm}
\SetAlgoLined

 \textbf{Input:} text,alpha,Lang \\
 sentences = split text into sentences\\
 \For{\textbf{each}  sentence $\in$ sentences  }{
    words = split sentence into words(text)\\
        change word to lowercase (words),\\
        Remove stop word(words, Lang)\\
    All words = append(words)
 }
\textbf{Output:} List of sentences, All words\\

\caption{pre-processing}
\label{alg:pre-processing}
\end{algorithm}

\subsection{Feature extracting}
Feature extraction consists of three steps: graph features extraction, textural features extraction and scoring these features. graph features and textural features extraction steps are parallel and extract features simultaneously. Each of the features represents different sight of the document, and the novelty of this paper is to fusion them to get an overview.

\subsubsection{Graph features}
Graphs are deployed to represent relations and make them easy to understand. Also, provide more computational results. Graphs consist of nodes and vertices, $G=\{V, E\}$, and nodes of the created unweighted and undirected graph are words, and vertices are 3-gram (trigram) of words. Figure \ref{fig:graph_example}  shows an example of a document and its graph. Once the graph is created, centralities are deployed to score each node. Graph centralities are measures of how nodes of a graph look from a different point of view. The centralities used in this paper are degree centrality(DE), closeness centrality(CL), betweenness centrality(BE), eigenvector centrality(EV), structural holes(SH), page rank(PR), clustering coefficient(CC), and eccentricity(EC). These eight centralities are the most used in NLP and graph representation scoring.

Several scores are assigned to every node based on their centrality, so eight scores are assigned to every node of the graph. In addition, this step should return a single score for every word; to that end, PCA is deployed. A PCA technique reduces the dimensionality of a dataset while increasing interpretability and minimizing information loss\cite{jolliffe_cadima_2016}. Nowadays, different versions of PCA with different aims are developed. PC1 is the first principal component used in NLP, and keyword extraction task\cite{vega2019multi} and we employed it to achieve one score per node (word). This progress is shown in algorithm \ref{alg:graph-features}.
   
\begin{minipage}{\textwidth}
    \centering
    \includegraphics[width=0.6\textwidth]{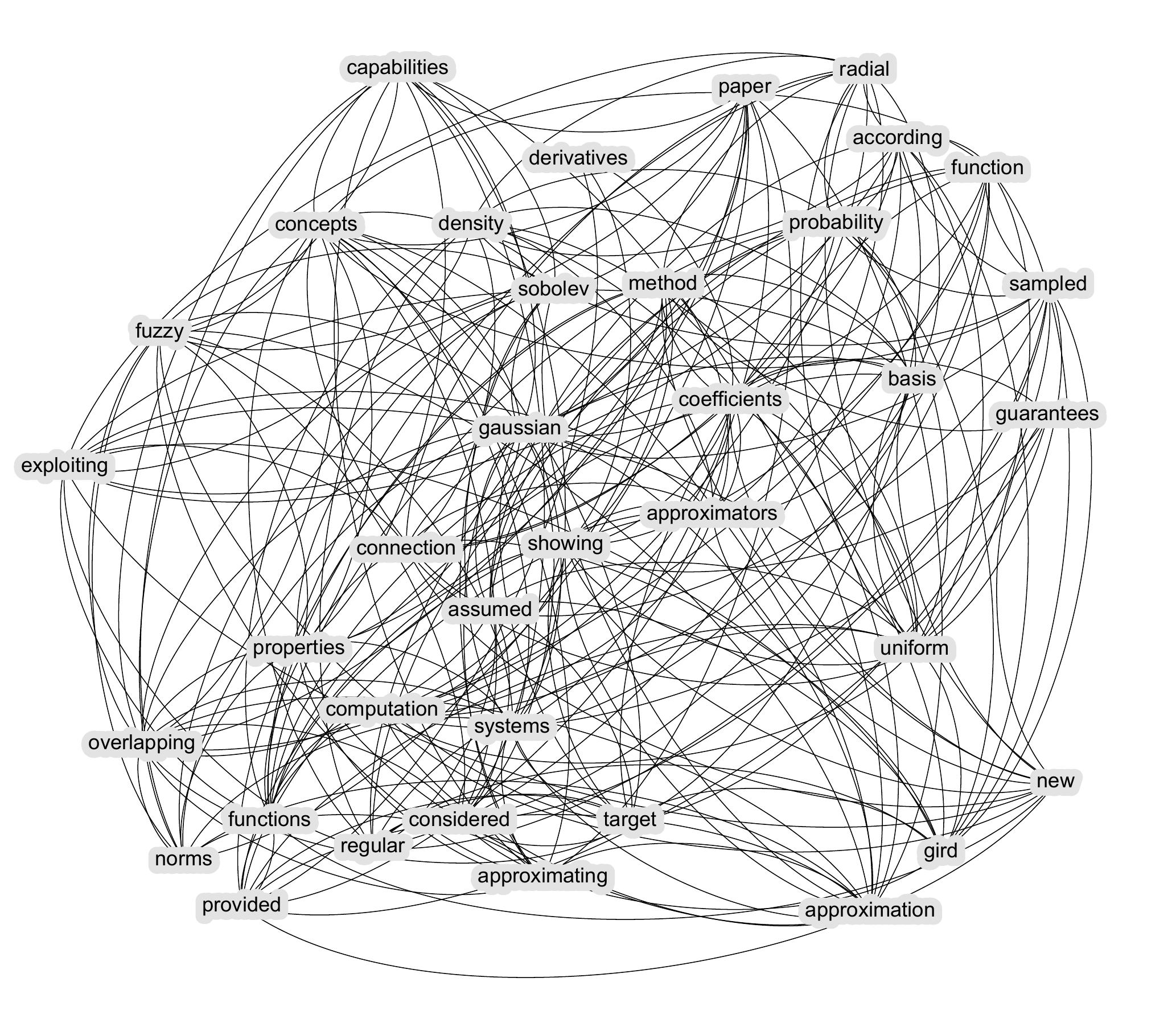}
    \captionof{figure}{The generated graph of example document fig \ref{fig:sample_1}.}
    \label{fig:graph_example}
\end{minipage}
    
\begin{algorithm}
 \SetAlgoLined
  \textbf{Input:} AllWords \\
\For {\textbf{each} $ word \in unique(AllWords)$}{
    G.node = unique(AllWords[word])\\
  }  
  \For {\textbf{each}  i $\in$ AllWords}{
    G.addEdge(AllWords list[i], AllWords list[i+2])\\
    G.addEdge(AllWords list[i], AllWords list[i+1])\\
    G.addEdge(AllWords list[i], AllWords list[i-1])\\
    G.addEdge(AllWords list[i], AllWords list[i-2])\\
    
  }  
\For {\textbf{each} Centrality $\in$ \{DE, CL, BE, EV, SH, PR, CC, EC\}}{

     \For {\textbf{each} word $\in$ words of Graph} {

    word centrality scores[Centrality]  =  calculate Centrality score (word,Centrality) \\
    
    }
    word.graph-Score =  Sum(word centrality score * PC1 coefficient for each centrality )
}
 \textbf{Output:} graph scores \\
 
 \caption{build graph and compute centrality measures Score}
 \label{alg:graph-features}
 \end{algorithm}
\subsubsection{textural features}
Word itself and its position in a document is essential and could make sense in understanding meaning. The features that have been utilized in this paper are capital letters, the position of the word, frequency of word in sentences of document, term frequency of word, and part of speech (POS) tag of word. Algorithm \ref{alg:local-features} details the features. Each of the features calculates a score. Finally, the textural feature score calculates in algorithm \ref{alg:local-features}.

\begin{algorithm}
    \SetAlgoLined
    \textbf{Input:} textural, words \\
    
    \For{\textbf{each} $word \in words$}{
        word.TCase = max (Letter case Term Frequency[word], Upper case Term Frequency[word] ) / (1 + ln (Term Frequency[word]))\\
        word.TPos = ln (3 + mean (offsets-sentences[word]))\\
        validTFs = word Term Frequency[words] \\
        avgTF = mean (validTFs)\\
        stdTF = stf (validTFs )\\
        word.TFNorm = Term Frequency[word] / ( avgTF + stdTF)\\
        word.TSent = length(offsets-sentences[word] /  length(sentences) \\
        word.Pos = part-of-speech(word) $->$ 'NN' = 1 , 'Adj' = 0.5 , 'V' = 0.25 \\
    
        word.sentence-Score = (word.TCase + word.TSent + word.TFNorm + word.Pos) / word.TPos\\
    }
    \textbf{Output:} textural scores\\
  
    \caption{Feature extraction and Compute textural measures scores}
    \label{alg:local-features}
\end{algorithm}
\subsection{Scores computation}
In this part, calculated scores of words by graph features and textural features are merged to claim one score per word. Multiply is used to combine two scores, as can be seen in algorithm \ref{alg:scores-computation}.

\begin{algorithm}
  \SetAlgoLined
  \textbf{Input:} graph-Score , textural-Score \\
  
  \For {\textbf{each} $word \in words$} {

    word.final-score = word.graph-Score × word.textural-Score\\
}

Sorting(words)\\
\textbf{Output:} final score
 
 \caption{compute final score from graph scores and Sentence scores}
 \label{alg:scores-computation}
\end{algorithm}

    \begin{table}[]
    \begin{center}
        \captionof{table}{Candidate keywords in couple with graph-based score, textual-based score and Final score of the example document.}
        \begin{tabular}{lccc}
            \hline
            \textbf{Words}  &\textbf{Graph-based score} &\textbf{Texture score}  &\textbf{Final score} \\
            \hline
            uniform                & 1.77   & 3.92    & 6.95  \\
            concepts               & 1.46   & 4.71    & 6.87  \\
            target                 & 1.61   & 3.93    & 6.34  \\
            systems                & 1.13   & 4.86    & 5.49  \\
            fuzzy                  & 1.11   & 4.94    & 5.48  \\
            overlapping            & 1.15   & 4.21    & 4.85  \\
            function               & 1.19   & 3.93    & 4.66  \\
            gaussian               & 1.14   & 3.91    & 4.45  \\
            properties             & 1.18   & 3.58    & 4.22  \\
            approximation          & 1.12   & 3.68    & 4.1   \\
            density                & 1.26   & 3.14    & 3.94  \\
            sobolev                & 1.12   & 3.51    & 3.92  \\
            norms                  & 1.1    & 3.46    & 3.81  \\
            paper                  & 1.1    & 3.41    & 3.75  \\
            capabilities           & 1.11   & 3.38    & 3.74  \\
            gird                   & 1.18   & 3.17    & 3.73  \\
            probability            & 1.18   & 3.14    & 3.7   \\
            guarantees             & 1.2    & 3.08    & 3.68  \\
            connection             & 1.15   & 3.12    & 3.6   \\
            sampledeither          & 1.12   & 3.19    & 3.57  \\
            coefficients           & 1.15   & 3.09    & 3.56  \\
            derivatives            & 1.16   & 3.07    & 3.55  \\
            radial                 & 1.11   & 3.12    & 3.48  \\
            computation            & 1.12   & 3.09    & 3.46  \\
            functionsapproximators & 1.11   & 3.11    & 3.44  \\
            basis                  & 1.09   & 3.11    & 3.4   \\
            method                 & 1.09   & 3.1     & 3.39  \\
            showing                & 1.25   & 2.7     & 3.38  \\
            considered             & 1.17   & 2.78    & 3.26  \\
            assumed                & 1.15   & 2.76    & 3.18  \\
            according              & 1.15   & 2.74    & 3.16  \\
            exploiting             & 1.16   & 2.72    & 3.16  \\
            provided               & 1.17   & 2.7     & 3.15  \\
            regular                & 1.12   & 2.53    & 2.84  \\
            new                    & 1.11   & 2.51    & 2.78 \\
            \hline
        \end{tabular}
     \end{center}
     \end{table}
\subsection{Key-phrase generation}
Key-phrase is a word or a couple of words (phrases) that convey the document's meaning. In the last part, keywords and scores are extracted. In this part, keywords are extracted from extracted text to generate key-phrases. N-grams is a general technique combining words to generate phrases, and determining N (length of N-gram) is one of the most issues. In this paper, HUPM\footnote{High Utility Pattern Mining}\cite{huang2015topic}, which core is FP-Growth, is applied to sentences to generate key-phrases. The phrase's score is calculated by summing the calculated scores for each keyword in the phrase.

    
        

    



            
            
    
        


  

\subsection{Keywords ranking}
The last step is to select key-phrases between candidates, which are generated from the previous parts. To this aim, the words and phrases are then sorted descendingly by their scores, and the top K is picked as keyphrases for the entire document.

    \begin{minipage}{\textwidth}
        \centering
        \includegraphics[width = 0.8 \textwidth]{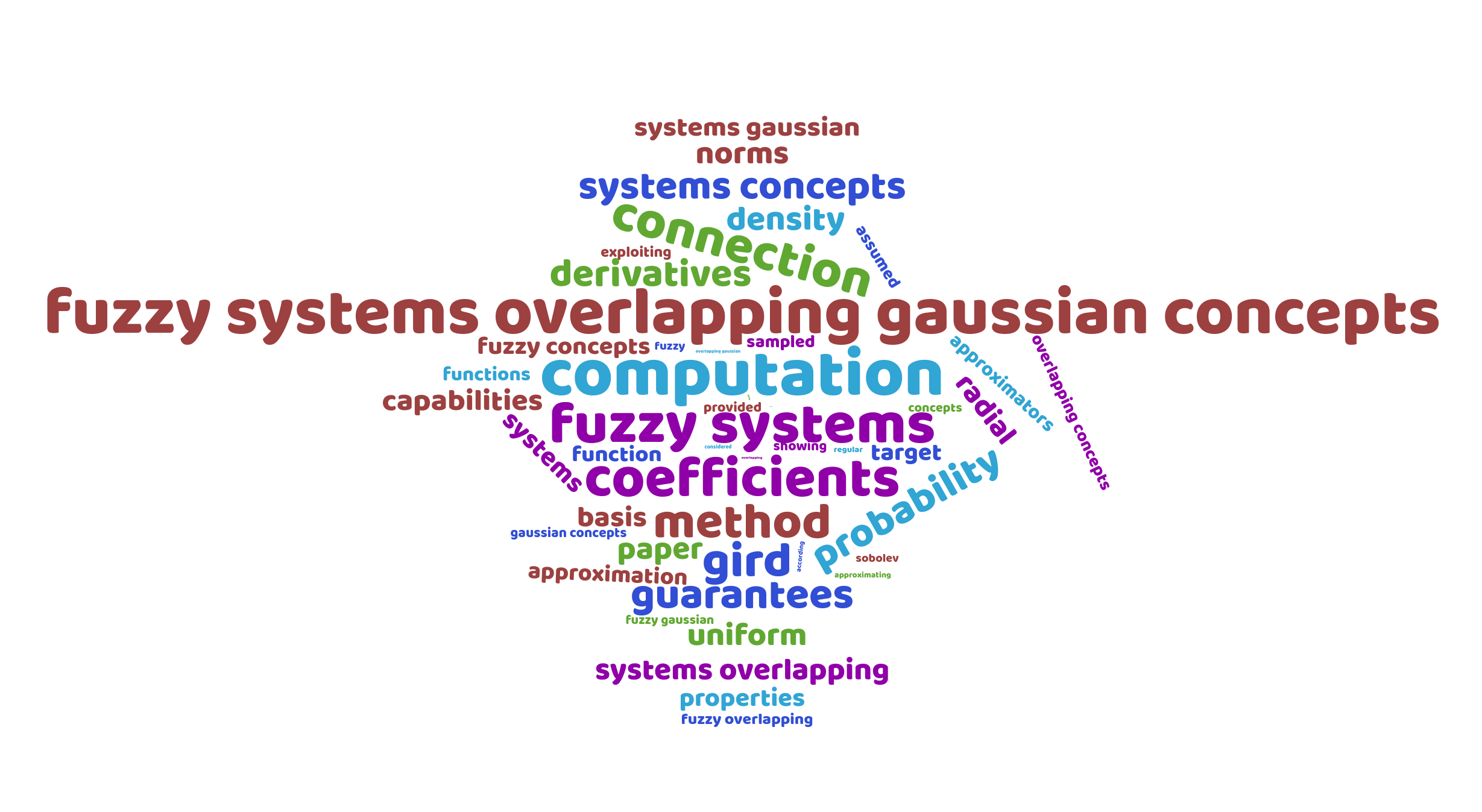}
        \captionof{figure}{The word cloud illustration of keywords and key-phrases example fig \ref{fig:sample_1}.}
        \label{fig:word-cloud}
    \end{minipage}

\section{Experiments and Results}   \label{sec:experiments_results}
In this section, the evaluation of the proposed methods and comparison are detailed. First, benchmark data sets are introduced, followed by performance metrics.

\subsection{Datasets}   \label{subsec:dataset}
Evaluation of the proposed methods has been done on 7 benchmark datasets. Summary of datasets is shown on table \ref{tbl:dataset_summary}. As shown in table \ref{tbl:dataset_summary}, various inputs and topics are selected for evaluation. To proving language-independency of the proposed method, two datasets are non-English.

\begin{table}[]
    \centering
    \caption{Dataset summary statistics}
    \label{tbl:dataset_summary}
    \begin{tabular}{cccccm{2cm}m{2cm}}
        \hline
        \textbf{Dataset}    &\textbf{Lang.}    &\textbf{Input} &\textbf{Topic} &\textbf{\# documents}    &\textbf{Avg \# words in documents}   &\textbf{\# keywords (percent)} \\
        \hline
        SemEval2010 \cite{Kim2013}  &en &Article    &Computer science   &243    &8332.34    &4002 (16.47)   \\
        \rowcolor{gray!10} SemEval2017 \cite{Augenstein2018}    &en &Paragraph  &Scientific &493    &178.22 &8969 (18.19)   \\
        Inspec \cite{Hulth2003} &en &Abstract   &Computer science   &2000   &128.20 &29230 (14.64) \\
        \rowcolor{gray!10} Fao30 \cite{Medelyan2006}    &en &Article    &Agricultural   &30 &4777.40    &997 (33.23)    \\
        Thesis100 \cite{medelyan_2015}  &en &Thesis &Scientific &100    &4728.86    &767 (7.67)\\
        \rowcolor{gray!10} pak2018 \cite{Campos2020}    &pl &Abstract   &Misc   &50 &97.36  &232 (4.64)  \\
        WikiNews \cite{Bougouin2013}    &fr &News   &Misc   &100   &293.52   &1177 (11.77)   \\
        \hline
    \end{tabular}
\end{table}

\subsection*{semeval2010}
Semival2010 consists of 244 articles indexed by ACM in four computer science area, namely; distributed systems, information retrieval, distributed artificial intelligence, and social science, introduced by\citet{Kim2013}. The input type is articles in the length of 6 till 8 pages and keywords are labelled by authors and expert. It is worthwhile to say that keywords may not be in the text. Summery of the dataset is shown in table \ref{tbl:dataset_summary}.

\subsection*{SemEval2017}
Semival2017 contains 500 articles abstract indexed by ScienceDirect equally divided in area of computer science, material engineering, and physics. Experts label the keywords. The dataset introduced fir the first time by\citet{Augenstein2018}. Summery of the dataset is shown in table \ref{tbl:dataset_summary}.

\subsection*{Inspec}
Inspec\cite{Hulth2003} includes 2000 articles abstracts in computer science collected between the years 1998 and 2002. Each document has two sets of keywords: the controlled keywords, which are manually controlled assigned keywords that appear in the Inspec thesaurus but may not appear in the document, and the uncontrolled keywords, which are freely assigned by the editors (that is, they are not restricted to the thesaurus or the document). In our experiments, we consider a union of both sets as the keywords. In table \ref{tbl:dataset_summary} summary of Inspec has been shown.

\subsection*{fao30}
fao30 is a collection of 30 agricultural documents from the Food and Agriculture Organization (FAO) of the United Nations. The fao30 is introduced in\cite{Medelyan2006}, and a summary is shown in table \ref{tbl:dataset_summary}. Six experts annotated foa30. fao30 is deployed to evaluate methods in long and non-scientific documents.

\subsection*{Thesis100}
Thesis100\cite{medelyan_2015} has included 100 master and PhD thesis of the University of Waikato, New Zeland in English. The domain of the thesis100 made available is quite different ranging from chemistry, computer science, philosophy, history, and others. Such as fao30, the Thesis100 is used to evaluating methods in long documents (more than 10 pages).

\subsection*{pak2018}
pak2018 is a dataset in Polish set of 50 abstracts of journals on technical topics collected from Measurement Automation and Monitoring\footnote{\url{http://pak.info.pl/}} and introduced by \citet{Campos2020}. The keywords are author-assigned, and a summary of the dataset is displayed in table \ref{tbl:dataset_summary}.

\subsection*{WikiNews}
WikiNews\cite{Bougouin2013} is a French corpus created from the French version of WikiNews\footnote{\url{https://www.wikinews.org/}} that contains 100 news articles published between May 2012 and December 2012 and manually annotated by at least three students. More details are given in table \ref{tbl:dataset_summary}.

\subsection{Results}    \label{subsec:results}

Calculating results starts with defining metrics. Afterward, metrics are defined, and the results of the proposed method compared to those of state-of-the-art methods are discussed.

\begin{table}[]
    \centering
    \caption{Comparison of features of the state-of-the-art methods and the proposed method.}
    \label{tbl:features-summary}
    \begin{tabular}{c|ccc|ccc}
        \hline
        \multirow{2}{*}{\textbf{Method}} &\multicolumn{3}{c|}{\textbf{Unsupervised}}   &\multicolumn{3}{c}{\textbf{Language dependence}}  \\
        &\textbf{Statistical}   &\textbf{Textual}   &\textbf{Graph-based}   &\textbf{Stop words}    &\textbf{POS tag}   &\textbf{Stream data}    \\
        \hline
        Proposed method & &\checkmark   &\checkmark   &\checkmark   &\checkmark   &   \\
        \rowcolor{gray!10} TF-IDF \cite{Lott2012}   &\checkmark &   &   &\checkmark &   &   \\
        KP-Miner \cite{El-Beltagy2009}  &\checkmark &   &   &\checkmark &   &   \\
        \rowcolor{gray!10} YAKE \cite{Campos2020}  &    &\checkmark    &   &\checkmark &   &   \\
        RaKUn \cite{rakun2019}   &   &\checkmark &   &   &   &   \\
        \rowcolor{gray!10} Text Rank \cite{Mihalcea2004}    &  &   &\checkmark &   &\checkmark & \\
        Single Rank \cite{Wan2008}  &    &   &\checkmark &\checkmark &\checkmark &   \\
        \rowcolor{gray!10} Topic Rank \cite{Bougouin2013}   &  &   &\checkmark &\checkmark &\checkmark &   \\
        Topical Page Rank \cite{Sterckx2015}    &    &   &\checkmark &\checkmark  &\checkmark &\checkmark   \\
        \rowcolor{gray!10} Posotion Rank \cite{Florescu2017}    &  &   &\checkmark &\checkmark  &\checkmark &   \\
        Multiparted Rank \cite{Boudin2018}  &    &   &\checkmark &\checkmark &\checkmark &\checkmark\\
        \rowcolor{gray!10} Expended Rank \cite{Wan2008} &  &   &\checkmark &   &\checkmark &   \\
        \hline
    \end{tabular}
\end{table}

Three primary metrics are used in evaluating keyword extraction methods are precision, recall, and F1-score. To calculating these metrics, 4 concepts, TP, TN, FP, FN, should be defined. TP denotes to the phrase correctly defined as a keyword. TN denotes to the phrase correctly define as a not keyword. FP denotes the phrase incorrectly define as a keyword, and FN denotes the phrase does not define as a keyword incorrectly. Respectively, precision and recall calculate by equation \ref{eq:precision} and equation \ref{eq:recall}. F1-score means the harmonic mean of the precision and recall, as equation \ref{eq:f1}.
Traditionally, keyword extraction is a ranking problem. Based on this, we opted to calculate Precision at k (Precision@k), Recall at k (Recall@k) and F1-score at k (F1-score@k) to determine the effectiveness of the proposed method.

\begin{equation}
    Precision = \frac{TP}{TP + FP}
\label{eq:precision}
\end{equation}

\begin{equation}
    Recall = \frac{TP}{TP + FN}
    \label{eq:recall}
\end{equation}

\begin{equation}
\label{eq:f1}
    F1-score = 2.\frac{Precision \times Recall}{Precision + Recall}
\end{equation}



The features of the state-of-the-art methods used as baselines compared with the proposed method have been shown in table \ref{tbl:features-summary}. As you see, two models of unsupervised, statistical and graph-based, methods with approaches on pre-processings detailed. To evaluation the state-of-the-art methods, implementation obtained from pke\footnote{python keyphrase extraction} \{\url{https://github.com/boudinfl/pke}\} developed by Boudin. Table \ref{tbl:result-precision}, \ref{tbl:result-recall}, \ref{tbl:result-f1} are the results of the methods in each of metrics.

Comparing the proposed method with other state-of-the-art methods, The proposed method performs the best in all metrics and datasets except the Thesis100. Thiesis100 is consists of very long texts, MSc thesis, with a low ratio of keywords to documents (7.67\% based on table \ref{tbl:dataset_summary}). Nevertheless, the proposed method obtains third place compared to other methods in all of the metrics. As seen, the graph-based methods outperformed this dataset. It is worthwhile to say that the comparisons have been made in Polish and France documents and the proposed method reaches the best results.

\begin{table}[]
    \centering
    \caption{Comparison of methods on Precision @ 10.}
    \label{tbl:result-precision}
    \begin{tabular}{cc|cccccccc}
        \hline
        \multicolumn{2}{c|}{\multirow{2}{*}{Methods}}    &\multicolumn{8}{c}{Dataset}   \\
        \cline{3-9}
        &   &SemEval10    &SemEval17    &Inspec &Fao30  &Thesis100 &pak18   &WikiNews   \\ 
        \hline
        \multicolumn{2}{c|}{Proposed Method (FRAKE)} &\textbf{0.415}   &\textbf{0.536} &\textbf{0.572}  &\textbf{0.294}  &0.24   &\textbf{0.126}    &\textbf{0.537} \\
        \cdashline{1-10}
        &TF-IDF \cite{Lott2012} &0.316  &0.488  &0.475  &0.251  &0.28   &0.104  &0.441   \\
        &KP-Miner \cite{El-Beltagy2009} &0.347  &0.398  &0.349  &0.222  &\textbf{0.291} &0.1    &0.452 \\
        &YAKE \cite{Campos2020}   &0.345    &0.334    &0.329    &0.1    &0.062  &0.054  &0.151  \\ 
        &Text Rank \cite{Mihalcea2004} &0.019  &0.335  &0.345  &0  &0.003  &0.008  &0.098 \\
        &Single Rank \cite{Wan2008}    &0.028   &0.151  &0.293  &0.007  &0.008  &0.027  &0.269 \\ 
        &Topic Rank \cite{Bougouin2013} &0.237  &0.436  &0.465 &0.129   &0.168  &0.04   &0.463 \\ 
        &Topical Page Rank \cite{Sterckx2015}  &0.027 &0.401 &0.42  &0.007 &0.11 &0.012  &0.331 \\
        &Position Rank \cite{Florescu2017}  &0.076 &0.438 &0.473 &0.025 &0.031 &0.035  &0.443 \\ 
        &MultiPartite Rank \cite{Boudin2018} &0.268 &0.458 &0.497 &0.159 &0.194 &0.037 &0.442 \\ 
        &Expanded Rank \cite{Wan2008}  & 0.027 & 0.396 & 0.413 & 0.007  &0.008 & 0.013  & 0.273 \\
        \hline
    \end{tabular}
\end{table}

\begin{table}[]
    \centering
    \caption{Comparison of methods on Recall @ 10.}
    \label{tbl:result-recall}
    \begin{tabular}{cc|cccccccc}
        \hline
        \multicolumn{2}{c|}{\multirow{2}{*}{Methods}}    &\multicolumn{8}{c}{Dataset}   \\
        \cline{3-9}
        &   &SemEval10    &SemEval17    &Inspec &Fao30  &Thesis100 &pak18   &WikiNews   \\
        \hline
        \multicolumn{2}{c|}{Proposed Method (FRAKE)} &\textbf{0.343}   &\textbf{0.544} &\textbf{0.607}  &\textbf{0.287}  &0.316   &\textbf{0.296} &\textbf{0.564} \\
        \cdashline{1-10}
        &TF-IDF \cite{Lott2012} &0.285  &0.5 &0.45   &0.226  &0.347  &0.247  &0.468 \\
        &KP-Miner \cite{El-Beltagy2009} &0.313  &0.405  &0.359  &0.2    &\textbf{0.354} &0.185  &0.464 \\
        &YAKE \cite{Campos2020}   &0.311    &0.342  &0.345  &0.1    &0.079  &0.123  &0.155  \\ 
        &Text Rank \cite{Mihalcea2004} &0.017  &0.31  &0.326 &0  &0.003 &0.17  &0.098 \\
        &Single Rank \cite{Wan2008}    &0.025   &0.141  &0.551 &0.006  &0.009  &0.013  &0.256 \\
        &Topic Rank \cite{Bougouin2013} &0.213  &0.400  &0.431  &0.116  &0.21   &0.08   &0.438  \\
        &Topical Page Rank \cite{Sterckx2015}  &0.024 &0.37 &0.395  &0.006  &0.012   &0.022 &0.315 \\
        &Position Rank \cite{Florescu2017}  &0.068 &0.405 &0.444 &0.023 &0.045  &0.073 &0.419 \\
        &MultiPartite Rank \cite{Boudin2018} &0.242 &0.421  &0.463  &0.143  &0.242  &0.074  &0.462  \\
        &Expanded Rank \cite{Wan2008}  &0.024   &0.366  &0.399  &0.006  &0.009  &0.027  &0.26 \\ 
        \hline
    \end{tabular}
\end{table}

\begin{table}[]
    \centering
    \caption{Comparison of methods on F1-score @ 10.}
    \label{tbl:result-f1}
    \begin{tabular}{cc|cccccccc}
        \hline
        \multicolumn{2}{c|}{\multirow{2}{*}{Methods}}    &\multicolumn{8}{c}{Dataset}   \\
        \cline{3-9}
        &   &SemEval10    &SemEval17    &Inspec &Fao30  &Thesis100 &pak18   &WikiNews   \\
        \hline
        \multicolumn{2}{c|}{Proposed Method (FRAKE)} &\textbf{0.375}   &\textbf{0.54} &\textbf{0.589}  &\textbf{0.29}  &0.272 &\textbf{0.177} &\textbf{0.55} \\
        \cdashline{1-10}
        &TF-IDF \cite{Lott2012} & 0.3 & 0.493 & 0.462 & 0.238 & 0.315&  0.146 & 0.454 \\
        &KP-Miner \cite{El-Beltagy2009} &0.329 &0.402 &0.354 & 0.21 &\textbf{0.319} &0.129  &0.457 \\
        &YAKE \cite{Campos2020}   &0.327    &0.338  &0.337  &0.1    &0.07   &0.075  &0.153   \\ 
        &Text Rank \cite{Mihalcea2004} &0.018   &0.322  &0.33 &0  &0.003    &0.011 &0.098 \\
        &Single Rank \cite{Wan2008}    &0.026   &0.145  &0.381  &0.007  &0.009  &0.017  &0.263 \\ 
        &Topic Rank \cite{Bougouin2013} &0.224  &0.417  &0.448  &0.122  &0.187  &0.053  &0.45 \\
        &Topical Page Rank \cite{Sterckx2015}   &0.026  &0.385  &0.407  &0.007  &0.012  &0.015 &0.323 \\
        &Position Rank \cite{Florescu2017}  &0.072  &0.421  &0.458  &0.024  &0.037  &0.047  &0.43  \\
        &MultiPartite Rank \cite{Boudin2018} &0.254 &0.439  &0.48   &0.15   &0.215  &0.05   &0.452 \\
        &Expanded Rank \cite{Wan2008}  &0.025   &0.381  &0.401  &0.007  &0.009  &0.017  &0.266 \\
        \hline
    \end{tabular}
\end{table}
\section{Conclusion}
A novel keyword extraction method called FRAKE is presented in this paper. FRAKE fuses two approaches, graph, and textural features, in order to extract keywords and key phrases. During the proposed method, five steps are included: pre-processing, extraction of graph features and textural features, computation of the Score, generation of key-phrases, and ranking of Key-phrases. It is shown that the proposed method performs best in English, Polish, and French texts.
\section*{Declarations}
\subsection*{Funding}
No funding was received to assist with the preparation of this manuscript.
\subsection*{Conflicts of interests}
The authors have no conflicts of interest to declare that are relevant to the content of this article.
\subsection*{Ethical approval}
This article does not contain any studies with human participants or animals performed by any of the authors.
\subsection*{Data availability}
Data sharing not applicable to this article as no datasets were generated or analysed during the current study.

\bibliography{mybibfile.bib}

\end{document}